\documentclass[letterpaper, 10 pt, conference]{ieeeconf}  % Comment this line out if you need a4paper

\usepackage{amsmath} % cmex10
\usepackage{amssymb}
\usepackage{siunitx}
\usepackage{caption}
\usepackage{graphicx}
\usepackage{float}
\usepackage{cite}
\usepackage{color}

\IEEEoverridecommandlockouts                              % This command is only needed if 
\newcommand{\ignore}[1]{}

\newcommand{\norm}[1]{\left\Vert#1\right\Vert} % Norm
\newcommand{\mc}[1]{\mathcal{#1}}  

\newcommand{\bbm}{\begin{bmatrix}}
\newcommand{\ebm}{\end{bmatrix}}
\newcommand{\bma}[1]{\left[\begin{array}{#1}}
\newcommand{\ema}{\end{array}\right]}

\DeclareMathAlphabet{\mbf}{OT1}{ptm}{b}{n}
\newcommand{\mbs}[1]{{\boldsymbol{#1}}}
\newcommand{\mbsdot}[1]{{\dot{\boldsymbol{#1}}}}

\newcommand{\mbfdot}[1]{{\dot{\mbf{#1}}}}
\newcommand{\mbfbar}[1]{{\bar{\mbf{#1}}}}
\newcommand{\mbfhat}[1]{{\hat{\mbf{#1}}}}

\def\fdotb{{\raisebox{-0.6ex}{ \kern0.2ex\raisebox{0.8ex}{\tiny $\hspace*{-1ex}\circ$}}}}
\def\fddotb{{\raisebox{-0.6ex}{ \kern0.2ex\raisebox{0.8ex}{\tiny $\hspace*{-1ex}\circ\circ$}}}}
\newcommand{\ura}[1]{{\underrightarrow{{#1}}}}

\newcommand{\trans}{{\ensuremath{\mathsf{T}}}} % transpose
\newcommand{\utimes}{ {\raisebox{-0.6ex}{ \kern-1.0ex\raisebox{0.6ex}{ \small $\mathsf{v}$}}} } % 
\newcommand{\beq}{\begin{equation}}
\newcommand{\eeq}{\end{equation}}
\newcommand{\bdis}{\begin{displaymath}}
\newcommand{\edis}{\end{displaymath}}
\newcommand{\beqarray}{\begin{eqnarray}}
\newcommand{\eeqarray}{\end{eqnarray}}
\newcommand{\beqarraynn}{\begin{eqnarray*}}
\newcommand{\eeqarraynn}{\end{eqnarray*}}

\renewcommand{\theenumii}{\arabic{enumii}}

\makeatletter
\renewcommand{\p@enumii}{\theenumi.}

\title{\LARGE \bf
Finite-Horizon LQR Control of Quadrotors on $SE_2(3)$
}

\author{Mitchell R. Cohen$^{1}$, Khairi Abdulrahim$^{2}$, and James Richard Forbes$^{3}$% <-this % stops a space
\thanks{This work was supported by the Mitacs Accelerate,  Universiti Sains Islam Malaysia (USIM), and the NSERC Discovery Grant program.}
\thanks{$^{1}$Mitchell R. Cohen ({\tt\small mitchell.cohen3@mail.mcgill.ca}) and $^{3}$James Richard Forbes ({\tt\small james.richard.forbes@mcgill.ca}) are with the Department of Engineering, McGill University, Montreal QC, Canada, H3A 0C3}
\thanks{$^{2}$Khairi Abdulrahim ({\tt\small khairiabdulrahim@usim.edu.my}) is with the Department of Electrical and Electronic Engineering, Universiti Sains Islam Malaysia, 71800, Nilai, Negeri Sembilan, Malaysia.}%
}

\begin{document}

%auto-ignore
% This is not a standalone latex document. To use this file
% as a cover page on an arXiv upload of a document that is 
% already accepted as some sort of IEEE publication, you must
%
%  1) add the following just after the \begin{document} line
%     of your main paper document
%
%         \input{arxiv-cover-ieee.tex}
%
%  2) and replace the relevant information in the block below.
%
% The relevant information has been parameterized as variables.
% Simply replace the variable values with your stuff and the 
% result should be good.
%
% Make sure to not include this file for ACTUAL submissions to 
% the IEEE. Luckily you can just comment in/out the 
% \input{arxiv-cover-ieee.tex} line.
%
% FYI: The exact citation with formatting can be obtained 
% from your paper's page on IEEE Xplore.
%
%%%%%%%%%%%%%%%%%%%%%%%%%%%%%%%%%%%%%%%%%%%%%%%%%%%%%%%%%%%%%%%
%%%%%%%%%%%%%%%%%%%%%% ADD YOUR INFO HERE %%%%%%%%%%%%%%%%%%%%%
%%%%%%%%%%%%%%%%%%%%%%%%%%%%%%%%%%%%%%%%%%%%%%%%%%%%%%%%%%%%%%%
\def \myJournal {IEEE Robotics and Automation Letters}
\def \myDoi {10.1109/LRA.2020.3010214}
\def \myPaperSiteName {IEEE Xplore}
\def \myPaperSiteLink {https://ieeexplore.ieee.org/document/9143426}
\def \myYear {2020}
\def \myPaperCitation{M. R. Cohen, K. Abdulrahim and J. R. Forbes, ``Finite-Horizon LQR Control of Quadrotors on $SE_2(3)$'' in \textit{IEEE Robotics and Automation Letters}, vol. 5, no. 4, pp. 5748-5755, Oct. 2020.}

%%%%%%%%%%%%%%%%%%%%%%%%%%%%%%%%%%%%%%%%%%%%%%%%%%%%%%%%%%%%%%%
%%%%%%%%%%%%%%%%%%%%%%%%%%%%%%%%%%%%%%%%%%%%%%%%%%%%%%%%%%%%%%%

\begin{figure*}[t]

\thispagestyle{empty}
\begin{center}
\begin{minipage}{6in}
\centering
This paper has been accepted for publication in \emph{\myJournal}. 
\vspace{1em}

This is the author's version of an article that has, or will be, published in this journal or conference. Changes were, or will be, made to this version by the publisher prior to publication.
\vspace{2em}

\begin{tabular}{rl}
DOI: & \myDoi\\
\myPaperSiteName: & \texttt{\myPaperSiteLink}
\end{tabular}

\vspace{2em}
Please cite this paper as:

\myPaperCitation

\vspace{15cm}
\copyright \myYear \hspace{4pt}IEEE. Personal use of this material is permitted. Permission from IEEE must be obtained for all other uses, in any current or future media, including reprinting/republishing this material for advertising or promotional purposes, creating new collective works, for resale or redistribution to servers or lists, or reuse of any copyrighted component of this work in other works.

\end{minipage}
\end{center}
\end{figure*}
\newpage
\clearpage
\pagenumbering{arabic} 

\maketitle
\thispagestyle{empty}
\pagestyle{empty}

%%%%%%%%%%%%%%%%%%%%%%%%%%%%%%%%%%%%%%%%%%%%%%%%%%%%%%%%%%%%%%%%%%%%%%%%%%%%%%%%
\begin{abstract}

This paper considers optimal control of a quadrotor unmanned aerial vehicles (UAV) using the discrete-time, finite-horizon, linear quadratic regulator (LQR). The state of a quadrotor UAV is represented as an element of the matrix Lie group of double direct isometries, $SE_2(3)$. The nonlinear system is linearized using a left-invariant error about a reference trajectory, leading to an optimal gain sequence that can be calculated offline. The reference trajectory is calculated using the differentially flat properties of the quadrotor. Monte-Carlo simulations demonstrate robustness of the proposed control scheme to parametric uncertainty, state-estimation error, and initial error. Additionally, when compared to an LQR controller that uses a conventional error definition, the proposed controller demonstrates better performance when initial errors are large.

\end{abstract}

%%%%%%%%%%%%%%%%%%%%%%%%%%%%%%%%%%%%%%%%%%%%%%%%%%%%%%%%%%%%%%%%%%%%%%%%%%%%%%%%
\section{INTRODUCTION}

Unmanned Aerial Vehicles (UAVs) are increasingly used for delivery, search and rescue, and inspection operations \cite{Shakhatreh2019, Chebrolu2018, vempati2018paintcopter, Nex2014}. Quadrotor UAVs are a popular option for tasks that require highly agile maneuvers in complex, constrained, environments \cite{Loquercio2019}. Control approaches that lead to robust and high performance operation of quadrotors are needed to execute tasks in a dependable manner.

The linear quadratic regulator (LQR) formulation is a popular means to design an optimal full-state-feedback controller for linear, or linearized, systems. In \cite{foehn2018onboard}, a quaternion-based LQR controller is presented where the nonlinear dynamics model of the quadrotor is linearized at the current timestep in order to synthesize the LQR gain. A similar approach is taken in \cite{Farrell2019}, but errors are defined in a multiplicative fashion for attitude and in an additive fashion for position and velocity. In both approaches, the collective thrust and angular velocity of the quadrotor are considered control inputs, necessitating a lower-level controller to regulate the true angular velocity to the desired angular velocity calculated by the LQR formulation. In addition, in \cite{Farrell2019}, the differentially flat properties of the dynamics of the quadrotor, as shown in \cite{mellinger2011minimum} and \cite{Faessler2018}, for example, are used to compute the reference trajectory and nominal control inputs. Rather than using an LQR approach in concert with linearization for control, nonlinear controllers applied to quadrotors are available in the literature \cite{lee2010geometric}, \cite{goodarzi2013geometric}.

% \colour{red}{Optimal control approaches based on linearization are contrasted by nonlinear controllers for quadrotor UAVs, with several notable examples being \cite{lee2010geometric} and \cite{goodarzi2013geometric}}.

% Optimal control for quadrotors has been considered in \cite{foehn2018onboard}, where an LQR controller is used to control both the rotational and translational dynamics of the quadrotor. The controller is quaternion-based, and relinearizes the system based on the current states at each control timestep to compute an optimal gain. A similar approach is employed in \cite{Farrell2019}, but this controller defines errors in a multiplicative fashion for attitude and in an additive fashion for position and velocity. In both controllers, the angular velocity of the quadrotor is considered as an intermediate control input, and a lower level controller regulates the true angular velocity to the desired angular velocity calculated by the LQR controller.

% A commonly known example of estimation-control duality is the duality between the Kalman filter and the linear-quadratic regulator \cite{todorov2008general}. 

The duality between full-state-feedback control and state estimation is well known. 
The formulation of the estimation problem for robotic and aerospace systems directly on matrix Lie groups has been considered within the invariant extended Kalman filter (IEKF) framework \cite{bonnable2009invariant, barrau2016invariant}. The use of the invariant framework has also been applied to the control problem in \cite{diemer2015invariant}, where an invariant LQG (ILQG) controller is derived for state estimation and control of a simplified car model. The ILQG controller derived in \cite{diemer2015invariant} linearizes the system about a desired reference trajectory. The formulation of the control problem directly on a matrix Lie group leads to improved robustness of the LQG controller to large initial deviations from the desired reference trajectory.

 The optimal control formulation proposed in this paper leverages key ideas from the invariant framework, much like \cite{diemer2015invariant} does. Rather than considering only the attitude as an element of a matrix Lie group, as is done in \cite{Farrell2019}, herein the quadrotor position, velocity, and attitude are all cast into an element of the group of double direct isometries, $SE_2(3)$, introduced in \cite{barrau2016invariant}. The error between the state and reference trajectory is defined as an element of $SE_2(3)$. Specifically, a left-invariant error definition is used. 
 The linearized system, once discritized, is used to compute an optimal gain sequence that can be calculated \emph{a priori}. Simulation results show that the use of the invariant framework leads to better performance when the error between the true states and reference states is large, which coincides with the findings of \cite{diemer2015invariant}.  
 
 % This paper combines ideas from \cite{foehn2018onboard}, \cite{Farrell2019}, and \cite{diemer2015invariant}, where LQR control of quadrotor UAVs is considered, as in \cite{foehn2018onboard} and  \cite{Farrell2019}, and ideas and insights from the invariant framework are used, as in \cite{diemer2015invariant}.

 % This paper builds on \cite{diemer2015invariant} by considering quadrotor UAV control
 
 % \cite{foehn2018onboard} and \cite{Farrell2019}
 
 % applying the key insights of the invariant framework to the control formulation of a quadrotor UAV.
 
 % \colour{blue}{What differentiates this paper from \cite{diemer2015invariant} is that herein, a discrete-time, finite-horizon optimal control problem is solved, a different matrix Lie group is used, and the invariant framework is adapted to the control formulation of a quadrotor UAV.}
 
 % because the Jacobians associated with the linearized system are less dependent on the reference trajectory, 
 
To account for unmodelled dynamics and parametric uncertainty, the proposed controller includes integral control similar to \cite{goodarzi2013geometric}. Feedforward terms, similar to \cite{Faessler2018}, are also used to account for the effects of drag forces and moments acting on the quadrotor.

In contrast to \cite{foehn2018onboard} where an infinite-horizon LQR control problem is solved at each timestep, in this paper, a finite-horizon LQR control problem is solved to compute a sequence of control gains. This allows a cost to be placed on the terminal state, which may be very useful if minimizing tracking error at the end of a given trajectory is important, such as during landing of the quadrotor.
%or the rendezvous of multiple quadrotors at a specific time to execute a cooperative task.

%  to obtain linearized error dynamics

The primary contribution of this paper is defining the error as an element of $SE_2(3)$, appropriately linearizing the nonlinear error dynamics, and synthesizing an optimal control gain sequence by solving the finite-horizon LQR control problem, all for quadrotor control. Doing so leads to improved robustness to initial error compared to when a conventional error definition is used. 
%This is similar to the conclusions drawn from \cite{diemer2015invariant} for the optimal control of a simplified car. 
Secondary contributions of this paper are the linearization about the reference trajectory leading to an optimal gain sequence that can be computed $\emph{a priori}$, and the inclusion of drag compensation terms and an integrator to account for parametric uncertainty within an optimal control formulation. This paper combines and builds on ideas from \cite{foehn2018onboard,Farrell2019,Faessler2018,diemer2015invariant}, where the specifics of quadrotor control is considered, as in \cite{foehn2018onboard,Farrell2019,Faessler2018}, and insights from the invariant framework are used to formulate the control, as in \cite{diemer2015invariant}.

The paper is organized as follows. Section~\ref{sec:preliminaries} outlines necessary preliminaries regarding matrix Lie groups, kinematics, and reference frames. The quadrotor equations of motion and details on the group $SE_2(3)$ are also provided. Section~\ref{sec:control} details the proposed controller, including the control objective and control inputs, as well as the feedback and feedforward components of the controller. Section~\ref{sec:simulation_results} provides simulation results as well as Monte-Carlo simulations, to demonstrate robustness of the proposed controller to state-estimation error, initial conditions, and parametric uncertainty.

\section{PRELIMINARIES} \label{sec:preliminaries}
\subsection{Matrix Lie Groups}
A general matrix Lie group $\mathcal{G}$ is composed of invertible $n \times n$ matrices with $k$ degrees of freedom closed under matrix multiplication \cite{hall2013lie}. The Lie algebra associated with $\mathcal{G}$, denoted $\mathfrak{g}$, is the tangent space of $\mathcal{G}$ at the identity element, denoted $\mbf{1}$. The exponential map is the mapping between the matrix Lie algebra and the matrix Lie group, such that $\mathrm{exp} (\cdot) : \mathfrak{g} \to \mathcal{G}$. For matrix Lie groups, the exponential map is the matrix exponential. The inverse of the exponential map is the logarithmic map and maps elements in the matrix Lie group to the matrix Lie algebra, such that $\mathrm{log} (\cdot) : \mathcal{G} \to \mathfrak{g}$. The ``vee" operator, $(\cdot)^{\vee} : \mathfrak{g} \to \mathbb{R}^k$, maps the matrix Lie algebra to a $k$ dimensional column matrix. The ``wedge" operator, $(\cdot)^{\wedge}$ is the inverse operator and is defined as $(\cdot)^\wedge : \mathbb{R}^k \to \mathfrak{g}$. An element of a matrix Lie group $\mathcal{G}$, $\mbf{X}$, can be perturbed using a left or right multiplication as $\mbf{X} = \mathrm{exp} \left(\delta \mbs{\xi}^\wedge \right) \mbfbar{X}$ or $ \mbf{X} = \mbfbar{X} \mathrm{exp} \left(\delta \mbs{\xi}^\wedge \right)$ respectively, where $\delta \mbs{\xi} \in \mathbb{R}^k$. For small $\delta \mbs{\xi}$, the element $\mathrm{exp} \left(\delta \mbs{\xi}^\wedge \right)$ can be linearized as $\mathrm{exp} \left(\delta \mbs{\xi}^\wedge \right) \approx \mbf{1} + \delta \mbs{\xi}^\wedge$.

\subsection{Kinematics and Reference Frames}
The inertial reference frame denoted $\mathcal{F}_a$ is composed of three orthonormal basis vectors \cite{hughes2012spacecraft}. An unforced particle in $\mathcal{F}_a$ is denoted $w$ \cite{bernstein2008newton}. The frame that rotates with the quadrotor body is denoted $\mathcal{F}_b$ and $\mbf{C}_{ab} \in SO(3)$ is the direction cosine matrix (DCM) that relates the attitude of $\mathcal{F}_a$ to the attitude of $\mathcal{F}_b$. The transpose of $\mbf{C}_{ab}$ is denoted $\mbf{C}_{ba}$ where $\mbf{C}_{ab} = \mbf{C}_{ba}^\trans$. A physical vector $\ura{v}$ can be resolved in either $\mathcal{F}_a$ or $\mathcal{F}_b$ as $\mbf{v}_a$ or $\mbf{v}_b$. The relation between the two is $\mbf{v}_a = \mbf{C}_{ab} \mbf{v}_b$. %  or $\mbf{v}_b = \mbf{C}_{ba} \mbf{v}_a$.

\subsection{Quadrotor Equations of Motion}
% \section{Quadrotor Equations of Motion} \label{sec:equations_of_motion}

Consider a quadrotor UAV modelled as a rigid body subject to gravitational, thrust, and drag forces, similar to \cite{Faessler2018}.
Denoting point $z$ to be the centre of mass of the vehicle, the translational and rotational dynamics are
	\begin{align} 
		\mbfdot{v}_a^{zw/a} & = \mbf{g}_a + \frac{\mbf{C}_{ab} \mbf{1}_3 f^\mathrm{T}}{m_\mathcal{B}} - \frac{1}{m_\mathcal{B}} \mbf{C}_{ab} \mbf{D} \mbf{C}_{ab}^\trans \mbf{v}_a^{zw/a}, \label{eq:equations_of_motion_trans} \\ 
		\mbsdot{\omega}_b^{ba} & = \mbf{J}_b^{\mathcal{B} z^{-1}} \left(\mbf{m}_b^{\mathrm{r}} - \mbs{\omega}_b^{ba^\times} \mbf{J}_b^{\mathcal{B} z} \mbs{\omega}_b^{ba} - \mbf{E} \mbf{C}_{ab}^{\trans} \mbf{v}_a^{zw/a} - \mbf{F} \mbs{\omega}_b^{ba}\right), \label{eq:equations_of_motion_rot}
	\end{align}
 where $m_\mathcal{B}$ is the mass of the UAV, $\mbf{J}_b^{\mathcal{B} z}$ is the second moment of mass of the quadrotor resolved in $\mathcal{F}_b$, $\mbf{g}_a = \begin{bmatrix} 0 & 0 & -g \end{bmatrix}^\trans$ is the gravity vector resolved in $\mc{F}_a$, $g = 9.81 \hspace{2mm} \mathrm{m}/\mathrm{s}^2$, $\mbs{\omega}_b^{ba}$ is the angular velocity of $\mathcal{F}_b$ relative to $\mathcal{F}_a$ resolved in $\mathcal{F}_b$, and $\mbf{v}_a^{zw/a}$ is the velocity of point $z$ relative to point $w$ with respect to $\mathcal{F}_a$, resolved in $\mathcal{F}_a$. The cross operator $(\cdot)^\times$ is a mapping from $\mathbb{R}^3$ to $\mathfrak{so}(3)$ such that $\mbf{a}^\times \mbf{b} = -\mbf{b}^\times \mbf{a} \hspace{2mm} \forall \mbf{a}, \mbf{b} \in \mathbb{R}^3$. The third column of the identity matrix is $\mbf{1}_3 = \begin{bmatrix} 0 & 0 & 1 \end{bmatrix}^\trans$. In addition, $f^\mathrm{T}$ represents the collective thrust force produced by the rotors, and $\mbf{m}_b^r$ represents a collective control moment produced by the rotors. The additional terms in \eqref{eq:equations_of_motion_trans} and \eqref{eq:equations_of_motion_rot} are drag terms that are linear in velocity and angular velocity. The constant diagonal matrix $\mbf{D} = \mathrm{diag} \left(d_x, d_y, d_z \right)$ is composed of rotor-drag coefficients, and $\mbf{E}$ and $\mbf{F}$ are constant drag matrices in the rotational dynamics. Identification of $\mbf{D}$, $\mbf{E}$, and $\mbf{F}$ is discussed in \cite{Faessler2018}.
 
 % The problem of identification of the matrices $\mbf{D}$, $\mbf{E}$, and $\mbf{F}$, is not considered in this paper, but more detail can be found \cite{Faessler2018}. In addition, adaptive mass estimation techniques for quadrotors can be found in \cite{bouadi2011adaptive} and \cite{fang2011adaptive}.
 
 % \cite{Faessler2018}

% The equations of motion are completely by 
The translational and rotational kinematics are given by 
    \begin{align} 
	    \mbfdot{r}_a^{zw} = \mbf{v}_a^{zw/a}, \hspace{5mm} \mbfdot{C}_{ab} = \mbf{C}_{ab} \mbs{\omega}_b^{ba^\times}, \label{eq:kin}
    \end{align}
where $\mbf{r}_a^{zw}$ is the position of $z$ relative to $w$ resolved in $\mc{F}_a$. Equations \eqref{eq:equations_of_motion_trans},  \eqref{eq:equations_of_motion_rot}, and \eqref{eq:kin} are the equations of motion. %  of the quadrotor.

% , that is, the inertial position of the quadrotor

\subsection{The Group of Double Direct Isometries}
% \section{The Group of Double Direct Isometries} \label{sec:se_23}
The quadrotor states $\mbf{C}_{ab}$, $\mbf{v}_a^{zw/a}$, and $\mbf{r}_a^{zw}$ can written as  an element of the group of double direct isometries, $SE_2(3)$, introduced in \cite{barrau2015non}. 
% Denoting $\mbf{X}$ as an element of $SE_2(3)$, the quadrotor states $\mbf{C}_{ab}$, $\mbf{v}_a^{zw/a}$, and $\mbf{r}_a^{zw}$ can be cast into $\mbf{X}$ as
Specifically, 
	\begin{align}
	        \mbf{X} = 
			\begin{bmatrix} 
				\mbf{C}_{ab} & \mbf{v}_a^{zw/a} & \mbf{r}_a^{zw} \\ 
				\mbf{0} & 1 & 0 \\ 
				\mbf{0} & 0 & 1
			 \end{bmatrix} \in SE_2(3).
	\end{align}
The column matrix $\mbs{\xi} \in \mathbb{R}^9$ can be mapped to the Lie algebra, $\mathfrak{se}_2(3)$, via  
    \begin{align}
        \mbs{\xi}^\wedge = 
            \begin{bmatrix} 
                \mbs{\xi}^\phi \\
                \mbs{\xi}^\mathrm{v} \\
                \mbs{\xi}^\mathrm{r}
            \end{bmatrix}^\wedge = 
            \begin{bmatrix} 
            \mbs{\xi}^{\phi^\times} & \mbs{\xi}^\mathrm{v} & \mbs{\xi}^\mathrm{r} \\ 
                            \mbf{0} & 0 & 0 \\
                            \mbf{0} & 0 & 0 
            \end{bmatrix} \in \mathfrak{se}_2(3).
    \end{align}
The exponential map from $\mathfrak{se}_2(3)$ to $SE_2(3)$ is given by 
    \begin{align}
        \mathrm{exp}\left(\mbs{\xi}^\wedge \right) = 
            \begin{bmatrix}     
                \mathrm{exp}_{SO(3)} \left(\mbs{\xi}^{\phi^\times} \right) & \mbf{J} \mbs{\xi}^\mathrm{v} & \mbf{J} \mbs{\xi}^\mathrm{r} \\
                \mbf{0} & 1 & 0 \\ 
                \mbf{0} & 0 & 1 
                \end{bmatrix} \in SE_2(3),
    \end{align}
where $\mbs{\xi}^\wedge \in \mathfrak{se}_2(3)$ and $\mbf{J}$ is given by 
    \begin{align}
        \mbf{J} = \frac{\sin \phi}{\phi} \mbf{1} + \left(1 - \frac{\sin \phi}{\phi} \right) \mbf{a} \mbf{a}^\trans + \frac{1 - \cos \phi}{\phi} \mbf{a}^\times,
    \end{align}
where $\phi = \norm{\mbs{\xi}^\phi}$ and $\mbf{a} = \mbs{\xi}^\phi / \phi$. In addition, the closed form solution for the exponential map from $\mathfrak{so}(3)$ to $SO(3)$, that being $\mathrm{exp}_{SO(3)} \left( \cdot \right)$, is given by the Rodrigues formula \cite{hughes2012spacecraft}.

\section{CONTROL} \label{sec:control}

\subsection{Control Objective}
% The control objective is to track a smooth trajectory on $SE_2(3)$. 
Denote the desired reference frame by $\mathcal{F}_r$, the desired velocity by $\mbf{v}_a^{z_\mathrm{r} w/a}$, and the desired position by $\mbf{r}_a^{z_\mathrm{r} w}$. The desired state can be cast into an element of $SE_2(3)$ as
	\begin{align}  \label{eq:reference_states}
		\mbf{X}^\mathrm{r} = 
			\begin{bmatrix} 
				\mbf{C}_{ar} & \mbf{v}_a^{z_\mathrm{r} w/a} & \mbf{r}_a^{z_\mathrm{r} w} \\ 
				\mbf{0} & 1 & 0 \\ 
				\mbf{0} & 0 & 1
			 \end{bmatrix} \in SE_2(3).
	\end{align}
For systems with states that are an element of a linear vector space, the tracking error, $\delta \mbf{x}$, is defined additively, as $\mbf{x} = \mbf{x}^\mathrm{r} + \delta \mbf{x}$ or $\delta \mbf{x} = \mbf{x} - \mbf{x}^\mathrm{r}$. However, for systems with states that are an element of a matrix Lie group, a multiplicative error definition is used. Consider the left-invariant tracking error \cite{barrau2016invariant}
	\begin{align} \label{eq:delta_X}
		\delta \mbf{X} = \mbf{X}^{-1} \mbf{X}^\mathrm{r} = 
		    \begin{bmatrix} 
		        \delta \mbf{C} & \delta \mbf{v} & \delta \mbf{r} \\
		        \mbf{0} & 1 & 0 \\
		        \mbf{0} & 0 & 1 
		      \end{bmatrix},
	\end{align}
where the individual errors $\delta \mbf{C}$, $\delta \mbf{v}$, and $\delta \mbf{r}$ are given by 
		\begin{align}
		\delta \mbf{C} & = \mbf{C}_{ab}^\trans \mbf{C}_{ar}, \label{eq:delta_C} \\
		\delta \mbf{v} & = \mbf{C}_{ab}^\trans \left(\mbf{v}_a^{z_\mathrm{r} w/a} - \mbf{v}_a^{zw/a} \right), \label{eq:delta_v} \\
		\delta \mbf{r} & = \mbf{C}_{ab}^\trans \left(\mbf{r}_a^{z_\mathrm{r} w} - \mbf{r}_a^{zw} \right). \label{eq:delta_r}
	\end{align}
The tracking error can also be written as
    \begin{align} \label{eq:left_invariant_error}
        \delta \mbf{X} = \mathrm{exp} \left(\delta \mbs{\xi}^\wedge \right).
    \end{align}
The control objective is to drive the tracking error to zero such that $\delta \mbf{X} = \mbf{1}$ or, equivalently, $\delta \mbs{\xi} = \mbf{0}$.

\subsection{Control Inputs}
As in \cite{foehn2018onboard} and \cite{Farrell2019}, it is assumed that a desired angular velocity can be accurately tracked using a low-level angular velocity controller. Under the assumption that the bandwidth of this lower-level controller is sufficiently high, the control inputs to the system are the collective thrust force, $f^\mathrm{T}$, and the body angular velocity, $\mbs{\omega}_b^{ba}$, such that $\mbf{u} = \begin{bmatrix} f^\mathrm{T} & \mbs{\omega}_b^{ba^\trans} \end{bmatrix}^\trans$.

\subsection{Control Structure Overview}

\begin{figure*}[tb]
   		\centering
      	\includegraphics[width=0.9\textwidth]{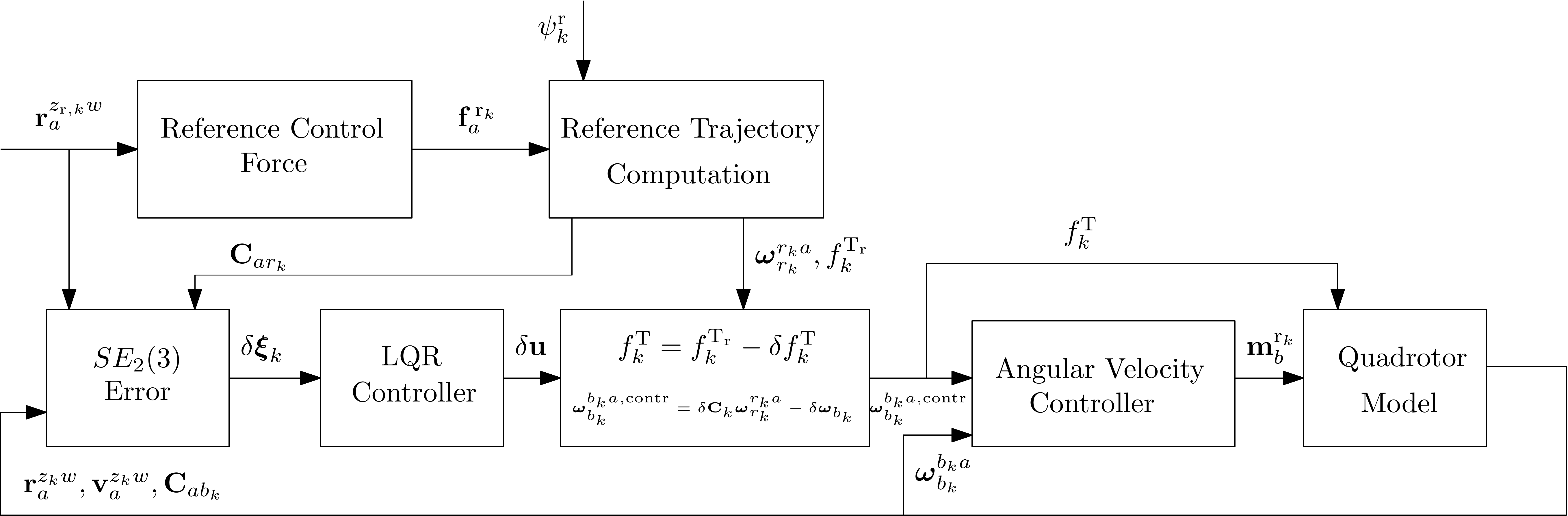}
		\caption{Proposed controller scheme.}
    	\label{fig:controller_scheme} 
  	\end{figure*}

An overview of the proposed control scheme is shown in Figure~\ref{fig:controller_scheme}. Specific aspects of the proposed control methodology are discussed in this section.

Recall that the dynamics of the quadrotor are \emph{differentially flat}. For differentially flat systems, a set of outputs can be found, equal to the number of inputs, such that all states and inputs can be determined from these outputs without integration \cite{murray1995differential} \cite{Fliess1995}. For a quadrotor, the differentially flat outputs are the position, $\mbf{r}_a^{zw}$, and yaw angle, $\psi$, of the vehicle. Consider a smooth trajectory in the flat outputs,
\begin{align} \label{eq:flat_outputs}
\mbs{\mu}(t) = \begin{bmatrix} \mbf{r}_a^{z_\mathrm{r} w^\trans}(t) & \psi^\mathrm{r}(t) \end{bmatrix}^\trans.
\end{align} 

The desired position and yaw angle at time $k$ are written $\mbf{r}_a^{z_{\mathrm{r}, k} w}$ and $\psi_k^\mathrm{r}$ respectively. From the reference position and yaw angle, the states $\mbf{C}_{a r_k}$, $\mbf{v}_a^{z_{\mathrm{r}, k} w/a}$, as well as feedforward reference thrust $f_k^{\mathrm{T}_{\mathrm{r}}}$ and reference angular velocity $\mbs{\omega}_{r_k}^{r_k a}$ can be found, as will be shown in Section~\ref{sec:reference_trajectory}. The reference attitude, velocity, and position at timestep $k$ are then placed into an element of $SE_2(3)$, as in \eqref{eq:reference_states}. Then, the left-invariant tracking error is computed as in \eqref{eq:left_invariant_error}, and the element $\delta \mbs{\xi}_k$ is extracted using $\delta \mbs{\xi}_k = \mathrm{log} \left(\delta \mbf{X}_k \right)^\vee$.

A feedback law of the form 
    \begin{align} \label{eq:state_feedback}
        \delta \mbf{u}_k = - \mbf{K}_{k, \mathrm{LQR}} \delta \mbs{\xi}_k 
    \end{align}
is used to find feedback control inputs $\delta \mbf{u} = \begin{bmatrix} \delta f_k^\mathrm{T} & \delta \mbs{\omega}_{b_k}^\trans \end{bmatrix}^\trans$, where $\mbf{K}_{k, \mathrm{LQR}}$ is the gain that is the solution to the discrete-time, finite-horizon LQR problem, discussed in Section \ref{sec:discrete_lqr}. In addition, $\delta \mbs{\omega}_{b_k}$ is the feedback desired angular velocity resolved in $\mathcal{F}_b$. The feedback control inputs are combined with the feedforward control inputs to compute the total control input $\mbf{u} = \begin{bmatrix} f_k^\mathrm{T} & \mbs{\omega}_{b_k}^{b_k a, \mathrm{contr}^\trans} \end{bmatrix}$, where
    \begin{align}
        f_k^\mathrm{T} & = f_k^{\mathrm{T}_\mathrm{r}} - \delta f_k^\mathrm{T}, \\
        \mbs{\omega}_{b_k}^{b_k a, \mathrm{contr}} & = \delta \mbf{C}_k \mbs{\omega}_{r_k}^{r_k a} - \delta \mbs{\omega}_{b_k}.
    \end{align}
Note that the reference angular velocity, $\mbs{\omega}_{r_k}^{r_k a}$, is resolved in $\mathcal{F}_r$, and therefore must be resolved in $\mathcal{F}_b$ using $\delta \mbf{C}_k$ before being combined with the feedback control input $\delta \mbs{\omega}_{b_k}$.
    
To determine the control torques to apply to the body, a controller with both feedforward and proportional-integral (PI) terms of the form
	\begin{align} \label{eq:attitude_controller}
		\mbf{m}_b^{\mathrm{r}_k} =
		\mbfhat{E} \mbf{C}_{a b_k}^\trans \mbf{v}_a^{z_k w/a} + \mbfhat{F} \mbs{\omega}_{b_k}^{b_k a} -\mbf{K}^\mbs{\omega} \mbf{e}_{b}^{\mbs{\omega}_k} - \mbf{K}^{i} \int_0^t \mbf{e}_{b}^{\mbs{\omega}_k} \mathrm{d} \tau,
	\end{align}
is used, where $\mbf{e}_b^{\omega_k} = \mbs{\omega}_{b_k}^{b_k a} - \mbs{\omega}_{b_k}^{b_k a, \mathrm{contr}}$ is the angular velocity error resolved in $\mathcal{F}_b$, $\mbfhat{E}$ and $\mbfhat{F}$ are estimates of the constant drag matricies $\mbf{E}$ and $\mbf{F}$, and $\mbf{K}^\omega = \mbf{K}^{\omega ^\trans} > \mbf{0}$ and $\mbf{K}^i = \mbf{K}^{i ^\trans} > \mbf{0}$ are PI control gains.

\subsection{Discrete-Time Finite-Horizon LQR} \label{sec:discrete_lqr}
Consider the discrete-time linearization of the system dynamics given by 
	\begin{align} 
		\delta \mbs{\xi}_{k+1} = \mbf{A}_{k} \delta \mbs{\xi}_{k} + \mbf{B}_{k} \delta \mbf{u}_k.
	\end{align}
Consider the cost function % \mbf{K}_{\mathrm{LQR}}
	\begin{align} 
		& J \left(\delta \mbf{u}_0,\ldots,\delta \mbf{u}_{N-1} \right) = \frac{1}{2} \delta \mbs{\xi}_N^\trans \mbf{S} \delta \mbs{\xi}_N^{\trans} \nonumber \\
		& \hspace{20pt} + \frac{1}{2} \sum_{k = 0}^{N-1} \left(\delta \mbs{\xi}_k^\trans \mbf{Q} \delta \mbs{\xi}_k + \delta \mbf{u}_k^\trans \mbf{R} \delta \mbf{u}_k \right),
	\end{align} 
 where $\mbf{S} = \mbf{S}^\trans \geq \mbf{0}$ is a weight on the terminal error, $\mbf{Q} = \mbf{Q}^\trans \geq \mbf{0}$ the error-deviation weight, and $\mbf{R} = \mbf{R}^\trans > \mbf{0}$ is the control effort weight.
 The solution to the discrete-time, finite horizon LQR problem is given by \cite{stengel1994optimal}
	\begin{align} 
		\delta \mbf{u}_k = - \mbf{K}_{k, \mathrm{LQR}} \delta \mbs{\xi}_k,
	\end{align}	
where $\mbf{K}_{k, \mathrm{LQR}}$ is the LQR gain at the $k$'th timestep, and the LQR gain is given by
		\begin{align} 
		\mbf{K}_{k, \mathrm{LQR}} = \mbfbar{R}_{k+1}^{-1} \mbf{B}_k^\trans \mbf{P}_{k+1} \mbf{A}_k,
	\end{align}
where $\mbfbar{R}_{k+1} = \mbf{R} + \mbf{B}_k^\trans \mbf{P}_{k+1} \mbf{B}_k$.
The sequence $\mbf{P}_k$, $k = 0, \ldots, N$ can be found by solving the discrete-time Riccati equation 
\begin{align} \label{eq:backwards_sweep}
		\mbf{P}_k = \mbf{A}_k^\trans \left(\mbf{P}_{k+1} - \mbf{P}_{k+1} \mbf{B}_k \mbfbar{R}_{k+1}^{-1} \mbf{B}_k^\trans \mbf{P}_{k+1} \right) \mbf{A}_k + \mbf{Q}
	\end{align}
backwards in time using the terminal condition $\mbf{P}_N = \mbf{S}$. The optimal gain sequence involves evaluating the Jacobians $\mbf{A}_k$ and $\mbf{B}_k$ at each controller timestep over the chosen horizon and solving \eqref{eq:backwards_sweep} backwards in time. The choice of horizon and timestep are user-defined parameters that depend on the chosen trajectory and the capabilities of the platform used.

% where $\mbf{R} = \mbf{R}^\trans > \mbf{0}$ is the control effort weight, $\mbf{Q} = \mbf{Q}^\trans \geq \mbf{0}$ the state deviation weight, and $\mbf{S} = \mbf{S}^\trans \geq \mbf{0}$ is the weight associated with the terminal cost.

\subsection{Linearization of Equations of Motion} \label{sec:linearization}
To find the linearized, discrete-time, equations of motion of the quadrotor, the equations of motion are linearized about the reference trajectory \cite{Zheng2012} in continuous-time, and then discritized using any desired discritization scheme. Note that the quadrotor rotational dynamics are not linearized, since $\mbs{\omega}_b^{ba}$ is not considered a state in $SE_2(3)$.

Using the $SE_2(3)$ error definitions in \eqref{eq:delta_X}, along with the angular velocity and thrust errors defined as
	\begin{align} 
		\delta \mbs{\omega}_b = \delta \mbf{C} \mbs{\omega}_r^{ra} - \mbs{\omega}_b^{ba}, \hspace{7mm} \delta f^{\mathrm{T}} = f^{\mathrm{T}_\mathrm{r}} -  f^{\mathrm{T}},
	\end{align}
the linearized dynamics of the system are given by
    \begin{align} \label{eq:linearized_dynamics}
        \delta \mbsdot{\xi} = \mbf{A} \delta \mbs{\xi} + \mbf{B} \delta \mbf{u},
    \end{align}
where the $\mbf{A}$ and $\mbf{B}$ matrices are given by \footnotesize
	\begin{align} \label{eq:jacobians}
	\mbf{A} = \begin{bmatrix}
				\mbf{0}  & \mbf{0} & \mbf{0} \\
				\mbf{A}_{2,1} & \mbf{A}_{2,2}  & \mbf{0} \\ 
				\mbf{0} & \mbf{1} & - \mbs{\omega}_r^{ra^\times}
			\end{bmatrix}, \hspace{2mm}
	\mbf{B} = \begin{bmatrix} 
				\mbf{0} & \mbf{1} \\ 
				\mbf{1}_3 \frac{1}{m_\mathcal{B}} & \mbf{0} \\
				\mbf{0} & \mbf{0}
			\end{bmatrix},
	\end{align}
\normalsize
where $\delta \mbs{\xi} = \begin{bmatrix} \delta \mbs{\xi}^{\phi^{\trans}} & \delta \mbs{\xi}^{\mathrm{v}^{\trans}} & \delta \mbs{\xi}^{\mathrm{r}^{\trans}} \end{bmatrix}^\trans$ and $\delta \mbf{u} = \begin{bmatrix} \delta f^\mathrm{T} & \delta \mbs{\omega}_b^\trans \end{bmatrix}^\trans$, and the individual elements $\mbf{A}_{2,1}$ and $\mbf{A}_{2,2}$ are given by
\footnotesize
    \begin{align} 
        \mbf{A}_{2,1} & = \frac{1}{m_\mathcal{B}} \left(\left(\mbf{D} \mbf{C}_{ar}^\trans \mbf{v}_a^{z_\mathrm{r} w/a} \right)^\times - \mbf{D} \left(\mbf{C}_{ar}^\trans \mbf{v}_a^{z_\mathrm{r} w/a} \right)^\times - \left(f^{\mathrm{T}_r} \mbf{1}_3\right)^\times \right), \\
            \mbf{A}_{2,2} & = -\mbs{\omega}_r^{ra^\times} - \frac{\mbf{D}}{m_\mathcal{B}}.
    \end{align}
\normalsize
A partial derivation of the linearized dynamics can be found in the Appendix. Unlike the LQR controllers presented in \cite{foehn2018onboard} and \cite{Farrell2019}, the proposed controller linearizes the dynamics about the reference states and inputs, and not the true states and inputs. This leads to Jacobians that only depend on the reference trajectory for the attitude $\mbf{C}_{ar}$, the angular velocity $\mbs{\omega}_r^{ra}$, velocity $\mbf{v}_a^{z_\mathrm{r} w/a}$, and thrust force $f^{\mathrm{T}_\mathrm{r}}$. Because the Jacobians can be calculated \emph{a priori}, the entire optimal gain sequence for a given trajectory can be calculated offline, which is computationally beneficial.

To reduce steady-state error due to unmodelled dynamics and parametric uncertainty, the system can then be augmented with integral control of the form
% an integrator. The integrator is
% The integrator used in the proposed controller is
	\begin{align}   \label{eq:integral_control}
		\mbs{\xi}^{i} = \int_0^{t} \left(c_1 \delta \mbf{r} + \delta \mbf{v} \right) \mathrm{d} \tau,
	\end{align}
where $c_1 > 0$ is a constant, $\delta \mbf{v}$ is  given in \eqref{eq:delta_v}, and $\delta \mbf{r}$  is given in \eqref{eq:delta_r}. This integrator has a similar form to the integrator in the outer-loop position controller of \cite{goodarzi2013geometric}.

The linearized dynamics can then be augmented with the integrator as
\begin{align} \label{eq:aug_jacobians}
		\underbrace{\begin{bmatrix} 
			\delta \mbsdot{\xi}^{\phi} \\
			\delta \mbsdot{\xi}^{\mathrm{v}} \\
			\delta \mbsdot{\xi}^{\mathrm{r}} \\
			\delta \mbsdot{\xi}^i
		\end{bmatrix}}_{\delta \mbsdot{\xi}^\mathrm{aug}}  = 
			\underbrace{\begin{bmatrix} 
				\mbf{0}  & \mbf{0} & \mbf{0} & \mbf{0} \\
				\mbf{A}_{2,1} & \mbf{A}_{2,2}  & \mbf{0} & \mbf{0} \\ 
				\mbf{0} & \mbf{1} & - \mbs{\omega}_r^{ra^\times} & \mbf{0} \\ 
				\mbf{0} & \mbf{1} & c_1 \mbf{1} & \mbf{0}
			\end{bmatrix}}_{\mbf{A}^{\mathrm{aug}}} 
			\underbrace{\begin{bmatrix}
				\delta \mbs{\xi}^{\phi} \\
			    \delta \mbs{\xi}^{\mathrm{v}} \\ 
			    \delta \mbs{\xi}^{\mathrm{r}} \\
				\delta \mbs{\xi}^i
			\end{bmatrix}}_{\delta \mbs{\xi}^\mathrm{aug}} +	
			\underbrace{\begin{bmatrix} 
                \mbf{B} \\ 
                \mbf{0}
			\end{bmatrix}}_{\mbf{B}^{\mathrm{aug}}} \delta \mbf{u}.
	\end{align}
The augmented continuous-time linearization of the system can be discritized using any desired discritization method to yield the discrete-time linear dynamics of the form
    \begin{align} 
        \delta \mbs{\xi}_{k+1}^{\mathrm{aug}} = \mbf{A}_k^{\mathrm{aug}} \delta \mbs{\xi}_k^\mathrm{aug} + \mbf{B}_k^\mathrm{aug} \delta \mbf{u}_k,
    \end{align}
where the augmented state-space and augmented Jacobians are used in the finite-horizon LQR gain computation. The system is discritized using the matrix exponential \cite{van1978computing}.

 \subsection{Computation of Reference Trajectory} \label{sec:reference_trajectory}
The reference trajectories for the states $\mbf{v}_a^{z_\mathrm{r} w/a}$, $\mbf{C}_{ar}$, and $\mbs{\omega}_r^{ra}$, and $f^{\mathrm{T}_\mathrm{r}}$ can be calculated from a given trajectory in the flat outputs, $\mbs{\mu}$. First, trajectories for the desired velocity $\mbf{v}_a^{z_\mathrm{r} w/a}$ and desired acceleration $\mbfdot{v}_a^{z_\mathrm{r} w/a}$ are found by differentiating the position trajectory. The reference velocity and acceleration at time $k$ are then denoted $\mbf{v}_a^{z_{\mathrm{r}, k} w/a}$ and $\mbfdot{v}_a^{z_{\mathrm{r}, k} w/a}$ respectively. These are used to find a reference control force through
	\begin{align} \label{eq:feedforward_control_force}
		\mbf{f}_{a}^{\, \mathrm{r}_k} = \hat{m}_{\mathcal{B}} \mbfdot{v}_a^{z_{\mathrm{r},k} w/a} + \hat{m}_{\mathcal{B}} g \mbf{1}_3 + \mbfbar{C}_{ar_k} \mbfhat{D} \mbfbar{C}_{ar_k}^\trans \mbf{v}_a^{z_{\mathrm{r},k} w/a},
	\end{align}
where $\mbfbar{C}_{ar_k}$ is initially set to $\mbf{0}$. In these feedforward terms, $\hat{m}_{\mathcal{B}}$ is the best estimate of the mass and $\mbfhat{D}$ is the best estimate of the diagonal drag coefficient matrix. 

Using both the reference control force $\mbf{f}_{a}^{\, \mathrm{r}_k}$ and the desired yaw angle $\psi_k^\mathrm{r}$, the reference DCM $\mbf{C}_{ar_k}$ can be computed. Denote the third basis vector of $\mathcal{F}_r$ as $\ura{r}^3$. Following \cite{mellinger2011minimum}, the components of $\ura{r}^3$ resolved $\mathcal{F}_a$ are given by
	\begin{align}
	\mbf{r}_a^{3_k} = \frac{\mbf{f}_{a}^{\,\mathrm{r}_k} }{\norm{\mbf{f}_{a}^{\,\mathrm{r}_k}}}.
	\end{align}
The components of an intermediate vector $\ura{c}^1$ resolved in inertial frame $\mathcal{F}_a$ are 
	\begin{align} 
		\mbf{c}_a^{1_k} = \begin{bmatrix} \cos(\psi_k^\mathrm{r}) & \sin(\psi_k^\mathrm{r}) & 0 \end{bmatrix}^\trans.
	\end{align}
The components of the remaining basis vectors defining reference frame $\mathcal{F}_r$ can then be found through
	\begin{align} 
		\mbf{r}_a^{2_k} = \frac{\mbf{r}_a^{{3_k}^\times} \mbf{c}_a^{1_k}}{\norm{\mbf{r}_a^{{3_k}^\times} \mbf{c}_a^{1_k}}}, \hspace{5mm}
		\mbf{r}_a^{1_k}  = \mbf{r}_a^{{2_k}^\times} \mbf{r}_a^{3_k}.
	\end{align}
The DCM relating the attitude of frame $\mathcal{F}_r$ relative to the attitude of the inertial frame $\mathcal{F}_a$ is  
	\begin{align} \label{eq:reference_attitude}
	\mbf{C}_{a r_k} = \begin{bmatrix} \mbf{r}_a^{1_k} & \mbf{r}_a^{2_k} & \mbf{r}_a^{3_k} \end{bmatrix}.
	\end{align}
Then, a new control force is computed using \eqref{eq:feedforward_control_force} setting $\mbfbar{C}_{ar_k} = \mbf{C}_{ar_k}$, and the procedure from~\eqref{eq:feedforward_control_force}-\eqref{eq:reference_attitude} is repeated until convergence of the reference control force. Upon convergence, the control force $\mbf{f}_a^{\, \mathrm{r}_k}$ is projected onto the reference $\ura{r}^3$ axis to yield the reference thrust force, $f_k^{\mathrm{T}_\mathrm{r}}$, given by $f_k^{\mathrm{T}_\mathrm{r}} = \mbf{1}_3^\trans \mbf{C}_{a r_k}^\trans \mbf{f}_{a}^{\, \mathrm{r}_k}$.
Next, the desired angular velocity is found using the discrete-time Poisson's equation
	\begin{align} 
		\mbf{C}_{a r_k} = \mbf{C}_{a r_{k-1}} \mathrm{exp}_{SO(3)} \left(T \mbs{\omega}_{r_k}^{r_k a^\times} \right),
	\end{align}
and solving for $\mbs{\omega}_{r_k}^{r_k a}$, where $T$ is the controller timestep.

\section{Simulation Results} \label{sec:simulation_results}
The proposed control scheme was tested in simulation on a model of a quadrotor with  equations of motion given by \eqref{eq:equations_of_motion_trans}. The mass, inertia, and drag properties of the quadrotor model, as well as the controller gains used in simulation, are given in Table~\ref{table:parameters}.

For all simulations performed, the same reference trajectory in the flat outputs is used, that being 
	\begin{align} \label{eq:trajectory}
		\mbf{r}_a^{z_\mathrm{r} w} (t) = \begin{bmatrix} 3 \sin(t) & 3 \cos(t) & 0.5 t \end{bmatrix}^\trans \mathrm{m}, \hspace{2mm} \psi^{\mathrm{r}}(t) = 0 \hspace{1mm} \mathrm{rad},
	\end{align}
where the desired position trajectory is a helix. More advanced methods of trajectory generation in the flat output space are discussed in \cite{mellinger2011minimum} and \cite{mellinger2012trajectory}.

% , but are not considered herein. 

% of the quadrotor, as discussed in \cite{mellinger2011minimum}, and  \cite{mellinger2012trajectory}, for example, are not considered in this paper.

The proposed $SE_2(3)$ LQR controller is compared to a conventional LQR controller that uses standard error definitions, similar to those defined in \cite{Farrell2019}. % The error definitions used by the conventional LQR controller are similar to those defined in \cite{Farrell2019}. 
The error definitions used by the conventional LQR control are
	\begin{align}  \label{eq:attitude_error_conventional}
		\delta \mbf{C} & = \mbf{C}_{ab}^\trans \mbf{C}_{ar}, \\ \label{eq:position_error}
		\delta \mbf{v}_a^{zw/a}  & = \mbf{v}_a^{z_\mathrm{r} w/a} - \mbf{v}_a^{zw/a}, \hspace{4mm}
		\delta \mbf{r}_a^{zw}  = \mbf{r}_a^{z_\mathrm{r} w} - \mbf{r}_a^{zw},
	\end{align}
leading to a linearization of the form
\begin{align} \label{eq:conventional_jacobians}
		\delta \mbfdot{x} = 
			\underbrace{\begin{bmatrix}
				\mbf{0} & \mbf{0} & \mbf{0} \\
			    \mbf{A}_{2,1}^\mathrm{c} & \mbf{A}_{2,2}^\mathrm{c} & \mbf{0} \\
				\mbf{0} & \mbf{1} & \mbf{0} \end{bmatrix}}_{\mbf{A}^{\mathrm{c}}}  \delta \mbf{x} + 
					\underbrace{\begin{bmatrix} 
						\mbf{0} & \mbf{1} \\
						\frac{\mbf{C}_{ar} \mbf{1}_3}{m_\mathcal{B}} & \mbf{0} \\
						\mbf{0} & \mbf{0}
					\end{bmatrix}}_{\mbf{B}^\mathrm{c}} \begin{bmatrix} \delta f^\mathrm{T} \\ \delta \mbs{\omega}_b \end{bmatrix},    
	\end{align}
where $\delta \mbf{x} = \begin{bmatrix} \delta \mbs{\xi}^{\phi^\trans} & \delta \mbf{v}_a^{zw/a^\trans} & \delta \mbf{r}_a^{zw^\trans} \end{bmatrix}^\trans$ and $\mbf{A}_{2,1}^\mathrm{c}$ and $\mbf{A}_{2,2}^\mathrm{c}$ are given by
    \begin{multline}
        \mbf{A}_{2,1}^\mathrm{c} = \frac{1}{m_\mathcal{B}} \biggl(\mbf{C}_{ar} \left(\mbf{D} \mbf{C}_{ar}^\trans \mbf{v}_a^{z_\mathrm{r} w/a} \right)^\times \\ - \mbf{C}_{ar} \mbf{D} \left(\mbf{C}_{ar}^\trans \mbf{v}_a^{z_\mathrm{r} w/a} \right)^\times  - \mbf{C}_{ar} \left(f^\mathrm{T}_r \mbf{1}_3 \right)^\times \biggr) ,
    \end{multline}
    \begin{align}
        \mbf{A}_{2,2}^\mathrm{c} = - \frac{1}{m_\mathcal{B}} \left(\mbf{C}_{ar} \mbf{D} \mbf{C}_{ar}^\trans \right) .
    \end{align}
Throughout, the superscript $(\cdot)^\mathrm{c}$ is used to denote the Jacobians associated with the conventional LQR controller.

%  in a similar way to the LQR controller that uses er $SE_2(3)$.
The conventional LQR controller is also augmented with an integrator in a similar way as the $SE_2(3)$ LQR controller is augmented with an integrator. The integrator used in the conventional controller is
	\begin{align}  
		\mbf{x}^{i} = \int_0^{t} \left(c_1 \delta \mbf{r}_a^{zw} + \delta \mbf{v}_a^{zw/a} \right) \mathrm{d} \tau,
	\end{align}
leading to a final row in the augmented Jacobians that is identical to that given in \eqref{eq:aug_jacobians}. %\eqref{eq:jacobians}.

% it is desired to make the linearized model as independent from the reference trajectory as possible. 

% Hence, modifying the Jacobians to render the Jacobians less depended on the reference trajectory is desirable.

In both the proposed $SE_2(3)$ LQR controller and the conventional LQR controller, linearization is performed about the reference trajectory. When the true quadrotor states are far from the reference trajectory, the linearization is no longer valid. The Jacobians $\mbf{A}$ and $\mbf{A}^\mathrm{c}$ given in \eqref{eq:jacobians} and \eqref{eq:conventional_jacobians}, respectively, depend more heavily on the reference trajectory through the terms $\mbf{C}_{ar}$ and $\mbf{v}_a^{z_\mathrm{r} w/a}$. By setting $\mbf{D} = \mbf{0}$, both $\mbf{A}$ and $\mbf{A}^\mathrm{c}$, respectively,  simplify to
\begin{align} \label{eq:simplified_linearization_se_23}
   \mbfbar{A} & = \begin{bmatrix}
				\mbf{0}  & \mbf{0} & \mbf{0} & \mbf{0} \\
				-\left(\frac{f^{\mathrm{T}_r} \mbf{1}_3}{m} \right)^\times & -\mbs{\omega}_r^{ra^\times}  & \mbf{0} & \mbf{0} \\ 
				\mbf{0} & \mbf{1} & - \mbs{\omega}_r^{ra^\times} & \mbf{0} \\ 
				\mbf{0} & \mbf{1} & c_1 \mbf{1} & \mbf{0}
			\end{bmatrix}, 
			\\ \label{eq:simplified_linearization_conventional}
	\mbfbar{A}^\mathrm{c} & = \begin{bmatrix}
				\mbf{0}  & \mbf{0} & \mbf{0} & \mbf{0} \\
			-\frac{\mbf{C}_{ar} \left(f^{\mathrm{T}_r} \mbf{1}_3 \right)^\times}{m_\mathcal{B}} & \mbf{0}  & \mbf{0} & \mbf{0} \\ 
				\mbf{0} & \mbf{1} & \mbf{0} & \mbf{0} \\ 
				\mbf{0} & \mbf{1} & c_1 \mbf{1} & \mbf{0}
			\end{bmatrix}.
	\end{align}
\normalsize
%
% Note that the drag terms in the linearization in both the Jacobians for the $SE_2(3)$ LQR controller given in \eqref{eq:jacobians}, and well as the Jacobians for the conventional controller given by \eqref{eq:conventional_jacobians}, make the Jacobians $\mbf{A}$ and $\mbf{A}^\mathrm{c}$ depend more heavily on the reference trajectory through the terms $\mbf{C}_{ar}$ and $\mbf{v}_a^{z_\mathrm{r} w/a}$. To reduce dependence of the Jacobians on the reference trajectory, drag can be neglected in the linearization. Denote the simplified Jacobians where drag has been neglected as $\mbfbar{A}$ and $\mbfbar{A}^\mathrm{c}$ for the $SE_2(3)$ controller and the conventional controller, respectively. These simplified Jacobians are written as
%
Now the Jacobian for the $SE_2(3)$ LQR controller, $\mbfbar{A}$, depends on the reference angular velocity, and the Jacobian for the conventional LQR controller, $\mbfbar{A}^\mathrm{c}$, depends on the reference attitude $\mbf{C}_{ar}$. Regardless of the inclusion of drag in the linearization, the Jacobian $\mbf{B}^\mathrm{c}$ is dependant on $\mbf{C}_{ar}$ while the Jacobian $\mbf{B}$ is not. Greater independence of the Jacobians $\mbfbar{A}$ and $\mbf{B}$ on the reference attitude is beneficial because the Jacobians are less sensitive to large attitude errors. Note that drag forces are still accounted for through the integrator, as well as the feedforward terms in the controller, through \eqref{eq:feedforward_control_force}.

% still valid when the true attitude of the quadrotor is far from the reference attitude.

% Note that in this simplified linearization, the Jacobian for the conventional controller, $\mbfbar{A}^\mathrm{c}$, depends on the reference attitude $\mbf{C}_{ar}$, while the Jacobian for the $SE_2(3)$ controller, $\mbfbar{A}$, only depends on the reference angular velocity. 

The performance of the proposed $SE_2(3)$ LQR controller and the conventional LQR controller are compared when using Jacobians that do and do not include drag. All parameters and control weights are kept constant between simulations. The robustness of the controllers to initial state error is first investigated, and then the robustness to parametric uncertainty is shown. Monte-Carlo simulation results are then presented to compare the performance of both controllers subject to state-estimation error, actuator dynamics, initial tracking error, and parametric uncertainty.

\subsection{Robustness to Initial Error}
Firstly, the robustness of both the $SE_2(3)$ LQR controller and the conventional LQR controller to initial error are investigated, for both the linearization including drag, yielding the Jacobians \eqref{eq:jacobians} and \eqref{eq:conventional_jacobians}, as well as linearization excluding drag, resulting in the Jacobians \eqref{eq:simplified_linearization_se_23} and \eqref{eq:simplified_linearization_conventional}.

\begin{figure}[h!]
		\centering
      	\includegraphics[width=0.9\linewidth]{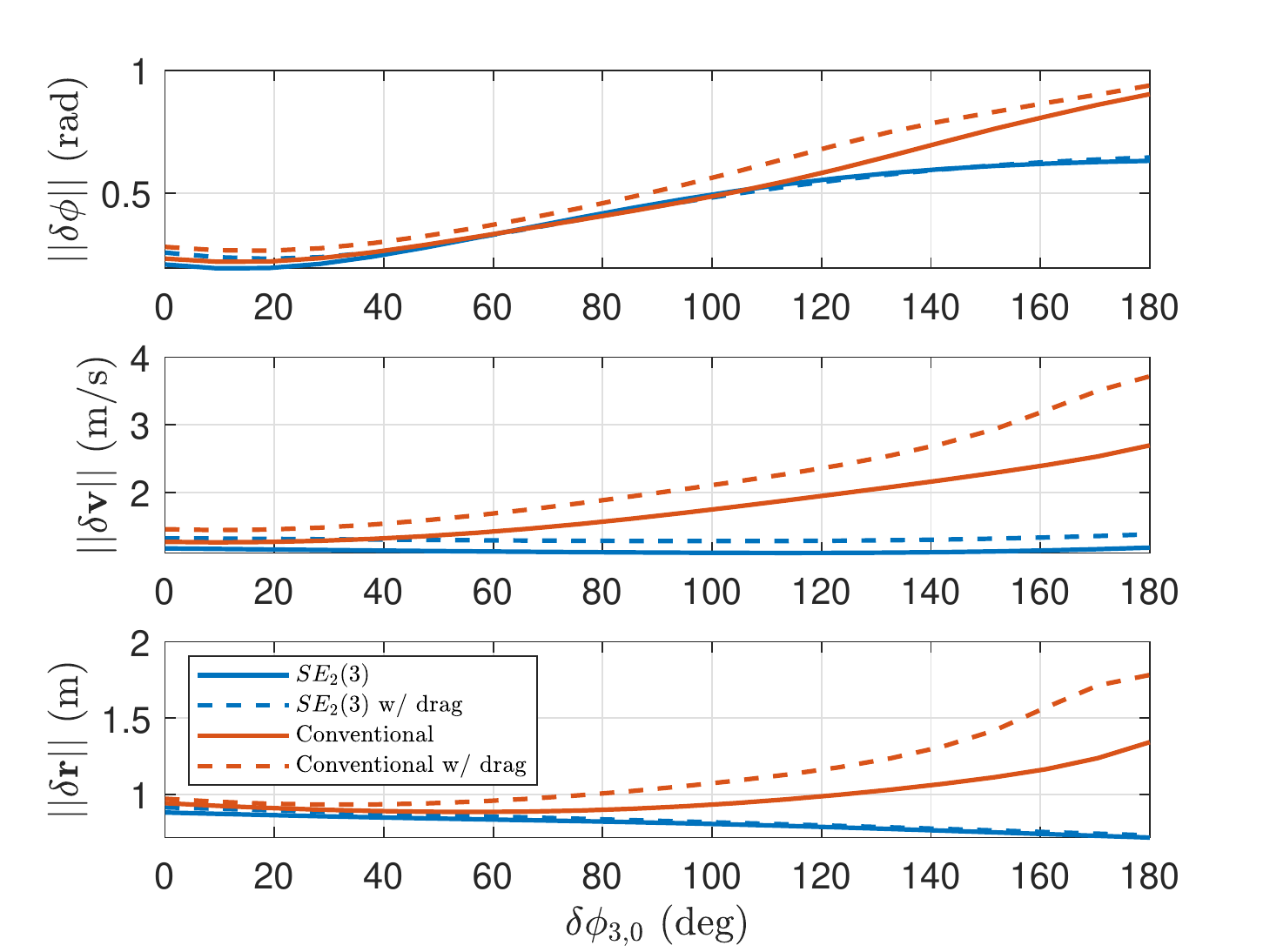}
		\caption{RMSE compared to initial heading error for both controllers.}
    	\label{fig:rmse_heading}%
\end{figure}

\begin{figure}[h!]
			\centering
      	\includegraphics[width=0.9\linewidth]{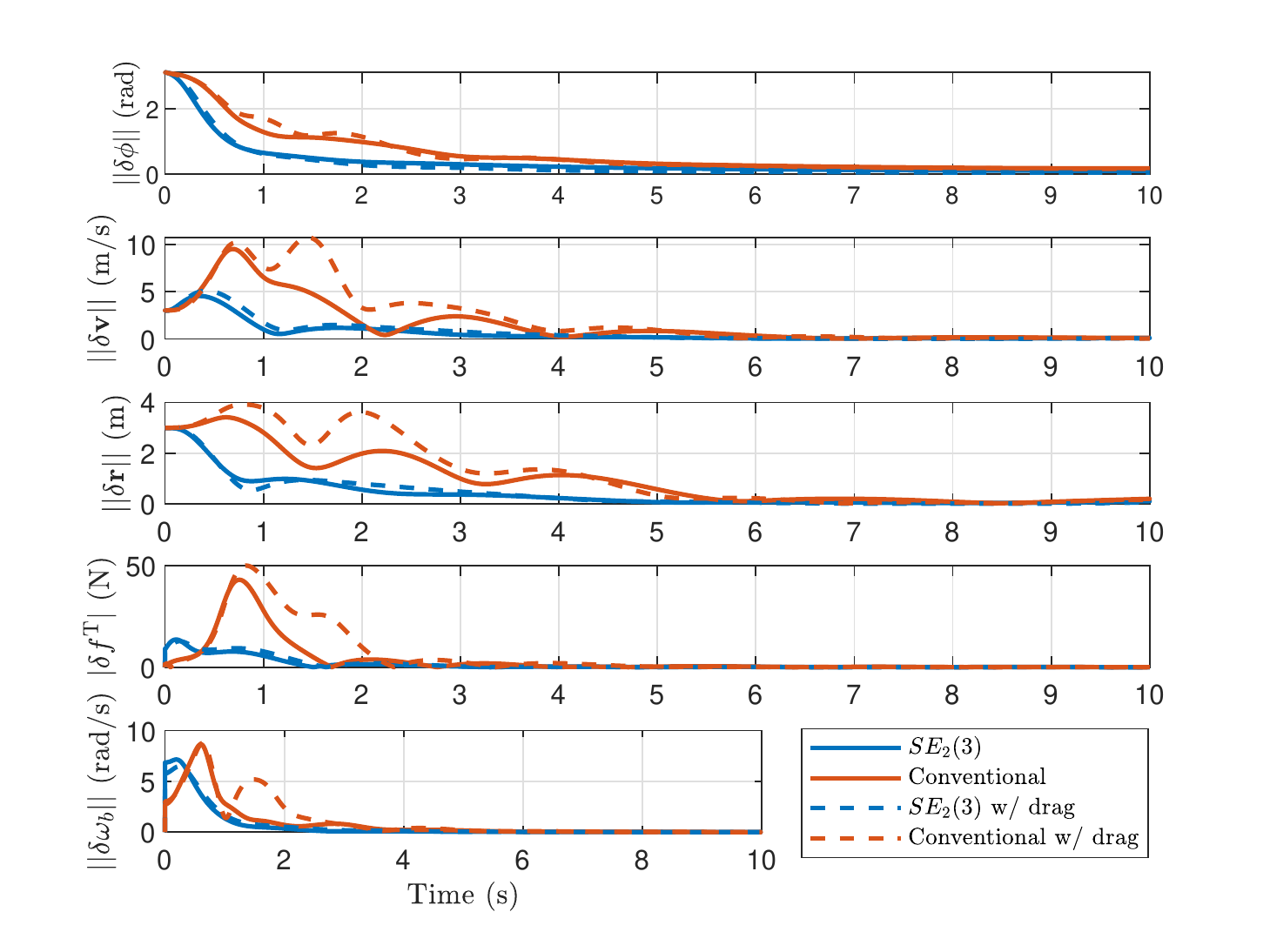}
      	\vspace{-5mm}
		\caption{Tracking results for $180^\circ$ initial heading error.}
    	\label{fig:180_degrees_heading_error}%
\end{figure}

For each controller, simulations are performed in a ``perfect" simulation environment, where the true states and parameters of the system are known exactly and are used in the controller. 
The RMSEs $\delta \mbs{\phi} = \mathrm{log}_{SO(3)} \left(\delta \mbf{C} \right)^\vee$, $\delta \mbf{v}$, and $\delta \mbf{r}$ for 10 seconds of simulation time, for various initial heading angle errors are shown in Figure~\ref{fig:rmse_heading}. The errors over time are also shown for 10 seconds of simulation in Figure~\ref{fig:180_degrees_heading_error}, when the initial heading error is $\delta \mbs{\phi}_{3,0} = 180^\circ$.

The $SE_2(3)$ LQR controller and the conventional LQR controller regulate the tracking errors to zero, when using the Jacobians that include and exclude  drag, as shown in Figure~\ref{fig:180_degrees_heading_error}. However, the Jacobians for %the LQR controller
the proposed $SE_2(3)$ LQR controller are not dependant on the reference attitude $\mbf{C}_{ar}$ when drag is neglected, and hence, improved performance is seen in the transient period where the actual quadrotor attitude is far from the desired attitude compared to each of the other three controllers tested. As the initial heading angle error increases, the proposed $SE_2(3)$ LQR controller outperforms the conventional LQR controller for both linearization schemes, as demonstrated in Figure~\ref{fig:rmse_heading}. This is consistent with the conclusions drawn from  \cite{diemer2015invariant}, where it was found that the use of the invariant error definition in the linearization lead to improved performance when the true states are very far from the reference states. In addition, these results are consistent with the fundamental results of the invariant framework when used for state estimation. As discussed in \cite{arsenault2019}, the performance benefit of the IEKF compared to a conventional EKF for state estimation can be attributed to better performance in the transient period. The linearization used in the IEKF depends less on the state estimate compared to the conventional EKF, and when the estimated states are far from the true states, the linearization using the invariant framework is more accurate.

\subsection{Robustness To Parametric Uncertainty}
\begin{figure}[h!]
			\centering
      	\includegraphics[width=0.93\linewidth]{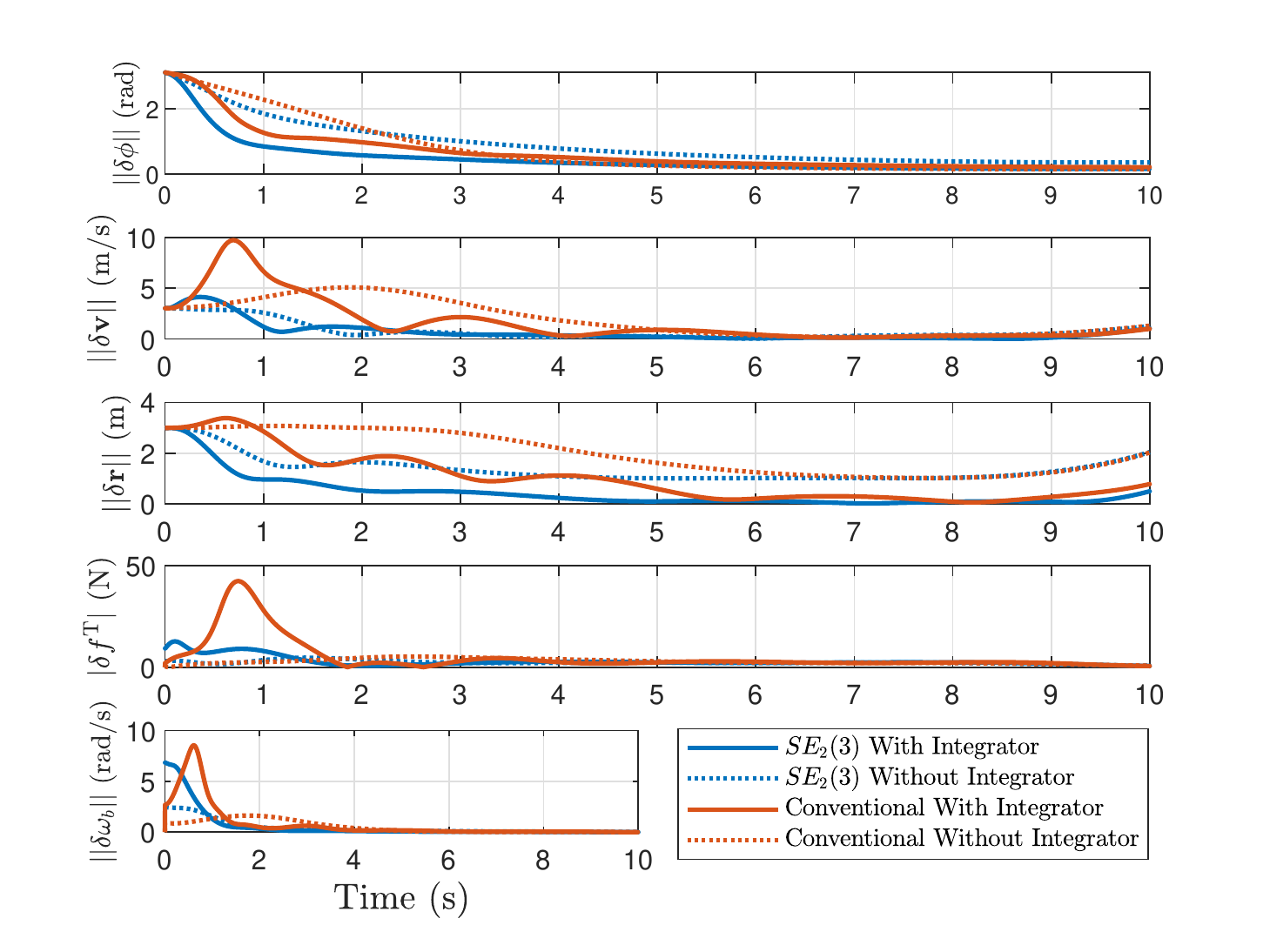}
		\caption{Tracking errors with and without integral control.}
    	\label{fig:integral_action}%
\end{figure}

To demonstrate robustness of the proposed $SE_2(3)$ LQR controller to parametric uncertainty, the simulation presented in Figure~\ref{fig:180_degrees_heading_error} is repeated, but this time, the estimated mass $\hat{m}_\mathcal{B}$ and estimated drag parameters $\mbfhat{E}$, $\mbfhat{F}$ and $\mbfhat{D}$ used in the controller are all set to 80\% of their true values. Figure~\ref{fig:integral_action} shows the tracking error and control effort for both controllers using the linearization scheme without drag, with and without integral control. Augmenting the controllers with an integrator of the form \eqref{eq:integral_control} reduces the steady state position error in the presence of parametric uncertainty. As shown in Figure~\ref{fig:integral_action}, for both the $SE_2(3)$ LQR and conventional LQR controllers, a steady-state error remains when integral control is not used. The particularly large steady-sate error in the position is due to the mass uncertainty. However, even without integral control, the performance of the $SE_2(3)$ LQR controller is improved in the transient period compared to the conventional LQR controller for large initial heading errors, as shown by the dotted lines in Figure~\ref{fig:integral_action}.

% Specifically, it was found that steady state error remains in the vertical position tracking without integral control, due to the uncertainty in the mass. 

\vspace{-3mm}
\subsection{Monte-Carlo Simulation Results}
Lastly, Monte-Carlo simulation results ensure robustness of the proposed controller to state-estimation error, actuator dynamics, initial error, and parametric uncertainty.

To ensure robustness to state-estimation error, an IEKF based on \cite{cohen2020navigation} is used to provide an estimate of the quadrotor states, which are an element of $SE_2(3)$, and sensor biases. Biased rate gyro and accelerometer measurements are used as interoceptive prediction sensors, and GPS position and velocity measurements, as well as magnetometer measurements, are used as correction sensors. The resultant state estimates are then used within the controller. 

To test the robustness of the proposed $SE_2(3)$ LQR controller and the conventional LQR controller to parametric uncertainty, the estimated mass, $\hat{m}_\mathcal{B}$, and estimated drag matrices $\mbfhat{E}$, $\mbfhat{F}$, and $\mbfhat{D}$ are all assumed to be perturbed from their true value such that
    \begin{align}
        \hat{m}_\mathcal{B} = \kappa_1 m_\mathcal{B}, \hspace{3mm} \mbfhat{E} = \kappa_2 \mbf{E}, \hspace{3mm} \mbfhat{F} = \kappa_3 \mbf{F}, \hspace{3mm}  \mbfhat{D} = \kappa_4 \mbf{D},
    \end{align}
where the parameters $\kappa_\imath$, $\imath = 1,2,3,4$, are taken from a normal distribution such that $\kappa_\imath \sim \mathcal{N} \left(1, \sigma_\imath^2 \right)$. The values for the parameters $\sigma_\imath$ are $\sigma_1 = 0.03 \mathbf{kg}$, and $\sigma_2 = \sigma_3 = \sigma_4 = 0.15$. For each trial, the initial position of the quadrotor, $\mbf{r}_a^{z_0 w}$ is randomized such that  $\mbf{r}_{a}^{z_0 w} \sim \mathcal{N} \left(0, \sigma_\mathrm{pos}^2 \right),$ where  $\sigma_\mathrm{pos}^2= 1 \hspace{1mm} \mathrm{m}$. The initial heading is also randomized such that the initial attitude of the quadrotor is set to $\mbf{C}_{ab_0} = \mbf{C}_3 \left(\psi_0 \right)$, where $\psi_0 \sim \mathcal{N} \left(0, \pi^2 \right)$. The control loop is run at 400 Hz and its outputs are all saturated at reasonable values. Actuator dynamics are modelled as a first-order low pass filter. Note that a control allocation problem can be solved to determine the required actuator inputs (i.e., the required motor speeds), to generate the collective desired thrust and desired control moments \cite{Johansen2013, marks2012control}.

\begin{figure}[h!]
			\centering
      	\includegraphics[width=0.9\linewidth]{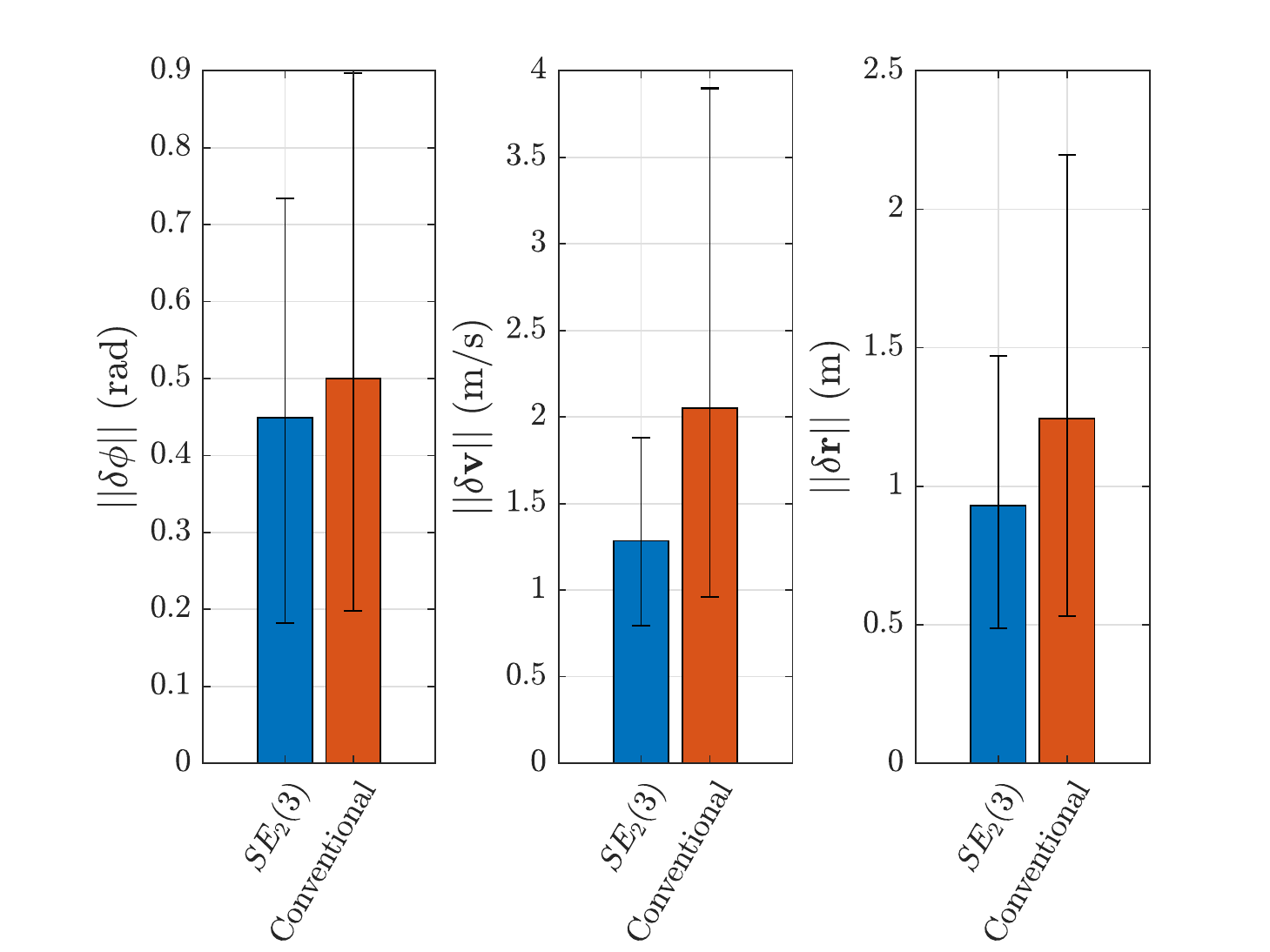}
		\caption{Monte-Carlo simulation results.}
    	\label{fig:monte_carlo_full}%
\end{figure}

The results for the average RMSEs for 100 Monte-Carlo trials for both controllers are shown in Figure~\ref{fig:monte_carlo_full}. The lower bounds of the error bars are set to a percentile of 2.5 and the upper bounds of the error bars are set to a percentile of 97.5, meaning the results of 95\% of the trials lie within the error bars. The proposed $SE_2(3)$ LQR controller outperforms the conventional LQR controller, even in the presence of state-estimation error, parametric uncertainty and unmodelled dynamics. It is noted that the performance benefit of the $SE_2(3)$ controller shown in the Monte-Carlo results is attributed to the improved robustness to initial error.

\begin{table}[h!] 
\footnotesize
 \caption{Quadrotor and Controller Parameter List}
 \centering
    \begin{tabular}{ |l | l |l  |}
    \hline
     Parameter & Value & Units\\ \hline
    $m_\mathcal{B}$ & 1.1 & \si{\kilogram} \\ \hline
    $\mbf{J}_b^{\mathcal{B} z} $ & diag(0.0112, 0.01123, 0.02108) & \si{\kilogram \meter\squared} \\ \hline
    $\mbf{D}$ & diag(0.605, 0.44, 0.275)  & \si{\per\second} \\ \hline
    $\mbf{E}$ & diag(0.05, 0.05, 0.05) & \si{\newton\per\second} \\ \hline   
    $\mbf{F}$ & diag(0.1, 0.1, 0.1) & \si{\newton\meter\second\per\radian} \\ \hline
    $\mbf{K}^\omega$ & diag(5, 5, 5) & \si{\newton\meter\second} \\ \hline
    $\mbf{K}^i$ & diag(3,3,3) & \si{\newton\meter\per\second} \\ \hline
    \end{tabular}
    \label{table:parameters}
\end{table}
\vspace{-2mm}
\section{Closing Remarks} % Conclusions and Future Work
In this paper, the discrete-time, finite-horizon LQR control problem using errors defined on $SE_2(3)$ was formulated and solved in order to control a quadrotor UAV. Based on the differentially flat properties of the quadrotor's dynamics, linearization was performed about the desired reference trajectory, leading to an \emph{offline} computation of optimal gain sequence for the controller. A feedforward controller and integrator were also included within the controller to account for drag, unmodelled dynamics, and parametric uncertainty. Simulation results demonstrate that the proposed controller scheme showed robustness to large initial heading error, parametric uncertainty, and state-estimation error. Future work will focus on an MPC approach, rather than an LQR approach, to control the UAV on $SE_2(3)$ in order to explicitly handle state and control constraints.

\appendices
\section{Quadrotor Dynamics Linearization Derivation}
Linearization of the quadrotor dynamics is partially presented in this Appendix. The attitude error propagation is given as
	\begin{align} \nonumber
		\delta \mbfdot{C} & = \mbfdot{C}_{ab}^{\trans} \mbf{C}_{ar} + \mbf{C}_{ab}^{\trans} \dot{\mbf{C}}_{ar} \\ \nonumber
		& = - \mbs{\omega}_b^{ba^\times} \mbf{C}_{ab}^\trans \mbf{C}_{ar} + \mbf{C}_{ab}^\trans \mbf{C}_{ar} \mbs{\omega}_r^{ra^\times} \\
		& = -\mbs{\omega}_b^{ba^\times} \delta \mbf{C} + \delta \mbf{C} \mbs{\omega}_r^{ra^\times}.
	\end{align}
Next, using $\mbs{\omega}_b^{ba} = -\delta \mbs{\omega}_b + \delta \mbf{C} \mbs{\omega}_r^{ra}$ yields
	\begin{align}
		\delta \mbfdot{C} = - \left(- \delta \mbs{\omega}_b + \delta \mbf{C} \mbs{\omega}_r^{ra} \right)^\times \delta \mbf{C} + \delta \mbf{C} \mbs{\omega}_r^{ra^\times}.
	\end{align}
Linearizing by letting $\delta \mbf{C} \approx \mbf{1} + \delta \mbs{\xi}^{\phi^\times}$ yields
\footnotesize
	\begin{multline}
		\delta \mbsdot{\xi}^{\phi^\times} = -\left(-\delta \mbs{\omega}_b + \left(\mbf{1} + \delta \mbs{\xi}^{\phi^\times} \right)  \mbs{\omega}_r^{ra} \right)^\times \left(\mbf{1} + \delta \mbs{\xi}^{\phi^\times} \right) \\
	    + \left(\mbf{1} + \delta \mbs{\xi}^{\phi^\times} \right) \mbs{\omega}_r^{ra^\times}  \\
	    \hspace{-5mm} = - \left(-\delta \mbs{\omega}_b + \mbs{\omega}_{r}^{ra} + \delta \mbs{\xi}^{\phi^\times} \mbs{\omega}_r^{ra} \right)^\times \left(\mbf{1} + \delta \mbs{\xi}^{\phi^\times} \right) \\ + \left(\mbf{1} + \delta \mbs{\xi}^{\phi^\times} \right) \mbs{\omega}_r^{ra^\times} \\ 
	     \hspace{10.5mm} = \delta \mbs{\omega}_b^\times + \delta \mbs{\omega}_b^\times \delta \mbs{\xi}^{\phi^\times} - \mbs{\omega}_r^{ra^\times} - \mbs{\omega}_r^{ra^\times} \delta \mbs{\xi}^{\phi^\times}
		 - \left(\delta \mbs{\xi}^{\phi^\times} \mbs{\omega}_r^{ra} \right)^\times \\ - \left(\delta \mbs{\xi}^{\phi^\times} \mbs{\omega}_r^{ra} \right)^\times \delta \mbs{\xi}^{\phi^\times} + \mbs{\omega}_r^{ra^\times} + \delta \mbs{\xi}^{\phi^\times} \mbs{\omega}_r^{ra^\times}.
	    \end{multline}
\normalsize
Neglecting higher order terms, and cancelling necessary terms, yields
	\begin{align}
		\delta \mbsdot{\xi}^{\phi^\times} = \delta \mbs{\xi}^{\phi^\times} \mbs{\omega}_r^{ra^\times} - \mbs{\omega}_r^{ra^\times} \delta \mbs{\xi}^{\phi^\times} - \left(\delta \mbs{\xi}^{\phi^\times} \mbs{\omega}_r^{ra} \right)^\times + \delta \mbs{\omega}_b^\times.
	\end{align}
Next, the identity
	$
		\mbf{u}^\times \mbf{v}^\times - \mbf{v}^\times \mbf{u}^\times = \left(\mbf{u}^\times \mbf{v} \right)^\times
	$
where $\mbf{u}$ and $\mbf{v} \in \mathbb{R}^3$, is used. Applying the identity yields
	\begin{align}
		\delta \mbsdot{\xi}^{\phi^\times} = -\left(\mbs{\omega}_r^{ra^\times} \delta \mbs{\xi}^{\phi} \right)^\times - \left(\delta \mbs{\xi}^{\phi^\times} \mbs{\omega}_r^{ra} \right)^\times + \delta \mbs{\omega}_b^{\times}.
	\end{align}
Uncrossing both sides yields
	\begin{align} \nonumber
		\delta \mbsdot{\xi}^\phi & = -\mbs{\omega}_r^{ra^\times} \delta \mbs{\xi}^\phi - \delta \mbs{\xi}^{\phi^\times} \mbs{\omega}_r^{ra} + \delta \mbs{\omega}_b
		= \delta \mbs{\omega}_b \nonumber
	\end{align}
In matrix form, the attitude dynamics are written
    \begin{align}
        \delta \mbsdot{\xi} = \begin{bmatrix} \mbf{0} & \mbf{0} & \mbf{0} \end{bmatrix} \delta \mbs{\xi} + \begin{bmatrix} \mbf{0} & \mbf{1} \end{bmatrix} \delta \mbf{u}.
    \end{align}

The error dynamics for $\delta \mbfdot{v}$, $\delta \mbfdot{r}$ and the integrator are derived in a similar fashion. The Jacobians for the conventional LQR controller are also derived similarly, using the conventional error definitions given in \eqref{eq:attitude_error_conventional}-\eqref{eq:position_error}.

\addtolength{\textheight}{-12cm}   % This command serves to balance the column lengths

%%%%%%%%%%%%%%%%%%%%%%%%%%%%%%%%%%%%%%%%%%%%%%%%%%%%%%%%%%%%%%%%%%%%%%%%%%%%%%%%

%%%%%%%%%%%%%%%%%%%%%%%%%%%%%%%%%%%%%%%%%%%%%%%%%%%%%%%%%%%%%%%%%%%%%%%%%%%%%%%%

%%%%%%%%%%%%%%%%%%%%%%%%%%%%%%%%%%%%%%%%%%%%%%%%%%%%%%%%%%%%%%%%%%%%%%%%%%%%%%%

\section*{ACKNOWLEDGMENT}

The authors graciously acknowledge funding from Mitacs Accelerate, Universiti Sains Islam Malaysia (USIM) and the National Science and Engineering Research Council (NSERC) of Canada.

\bibliographystyle{IEEEtran}
\bibliography{ref}

\end{document}